\documentclass[final]{cvpr}

\usepackage{times}
\usepackage{epsfig}
\usepackage{graphicx}
\usepackage{amsmath}
\usepackage{amssymb}
\usepackage{graphicx}
\usepackage{tabularx}
\usepackage{amsthm}
\newcolumntype{Y}{>{\centering\arraybackslash}X}
\usepackage{booktabs}
\usepackage{multirow}
\usepackage{float} 
\usepackage{subfigure}
\usepackage[ruled,linesnumbered]{algorithm2e}

\usepackage[pagebackref=true,breaklinks=true,colorlinks,bookmarks=false]{hyperref}

\def\cvprPaperID{4743} 
\def\confYear{CVPR 2021}

\begin{document}

\title{V$^2$L: Leveraging Vision and Vision-language Models into Large-scale Product Retrieval}

\author{Wenhao Wang$^{*}$, Yifan Sun$^{*}$, Zongxin Yang$^{\dagger}$, Yi Yang$^{\dagger}$\\
$^{*}$Baidu Research, $^{\dagger}$Zhejiang University\\
{\tt\small wangwenhao0716@gmail.com, sunyifan01@baidu.com, yangzongxin@zju.edu.cn, yee.i.yang@gmail.com}
}

\maketitle

\begin{abstract}
Product retrieval is of great importance in the ecommerce domain. This paper introduces our 1st-place solution in eBay eProduct Visual Search Challenge (FGVC9), which is featured for an ensemble of about 20 models from vision models and vision-language models. While model ensemble is common, we show that combining the vision models and vision-language models brings particular benefits from their complementarity and is a key factor to our superiority. Specifically, for the vision models, we use a two-stage training pipeline which first learns from the coarse labels provided in the training set and then conducts fine-grained self-supervised training, yielding a coarse-to-fine metric learning manner. For the vision-language models, we use the textual description of the training image as the supervision signals for fine-tuning the image-encoder (feature extractor). With these designs, our solution achieves 0.7623 MAR@10, ranking the first place among all the competitors. The code is available at: \href{https://github.com/WangWenhao0716/V2L}{V$^2$L}.

\vspace*{-4mm}

\end{abstract}

\section{Introduction}
Given a query image, the goal of product retrieval is to determine whether there are the same product images from a reference dataset. It plays an important role in the e-commerce domain. There are two main challenges. First, because product retrieval is an instance of super fine-grained recognition, it is very difficult for algorithms to distinguish two products with subtle visual differences. Second, building large-scale product training datasets with fine-grained level labels is very time-consuming and expensive.
 \par

This paper tries to leverage \textbf{V}ision and \textbf{V}ision-\textbf{L}anguage models (V$^2$L) to compete for eBay eProduct Visual Search Challenge (FGVC9) \cite{yuan2021eproduct} at CVPR'22. This competition builds a dataset with $2.5$ million product images. Different from other similar datasets \cite{bai2020products,Le_2020_ECCV}, the dataset features: (1) It does not have fine-grained labels, \textit{i.e.} it only provides categorical labels rather than product labels. In this way, we can build self-supervised learning algorithms on our previous winning solutions \cite{wang2021d, wang2021bag}. (2) It provides text information (title) for each product image in the training set. In this way, we can explore multimodal learning methods. To explore the highest performance, we do not limit the amount of GPU and CPU resources. We use $6$ standard Nvidia $8$-A100 GPU servers and about $100$ Intel 6240 CPUs (with $36$ cores per CPU) in the competition. \par
\begin{figure}[t]
\centering 
\includegraphics[width=0.47\textwidth]{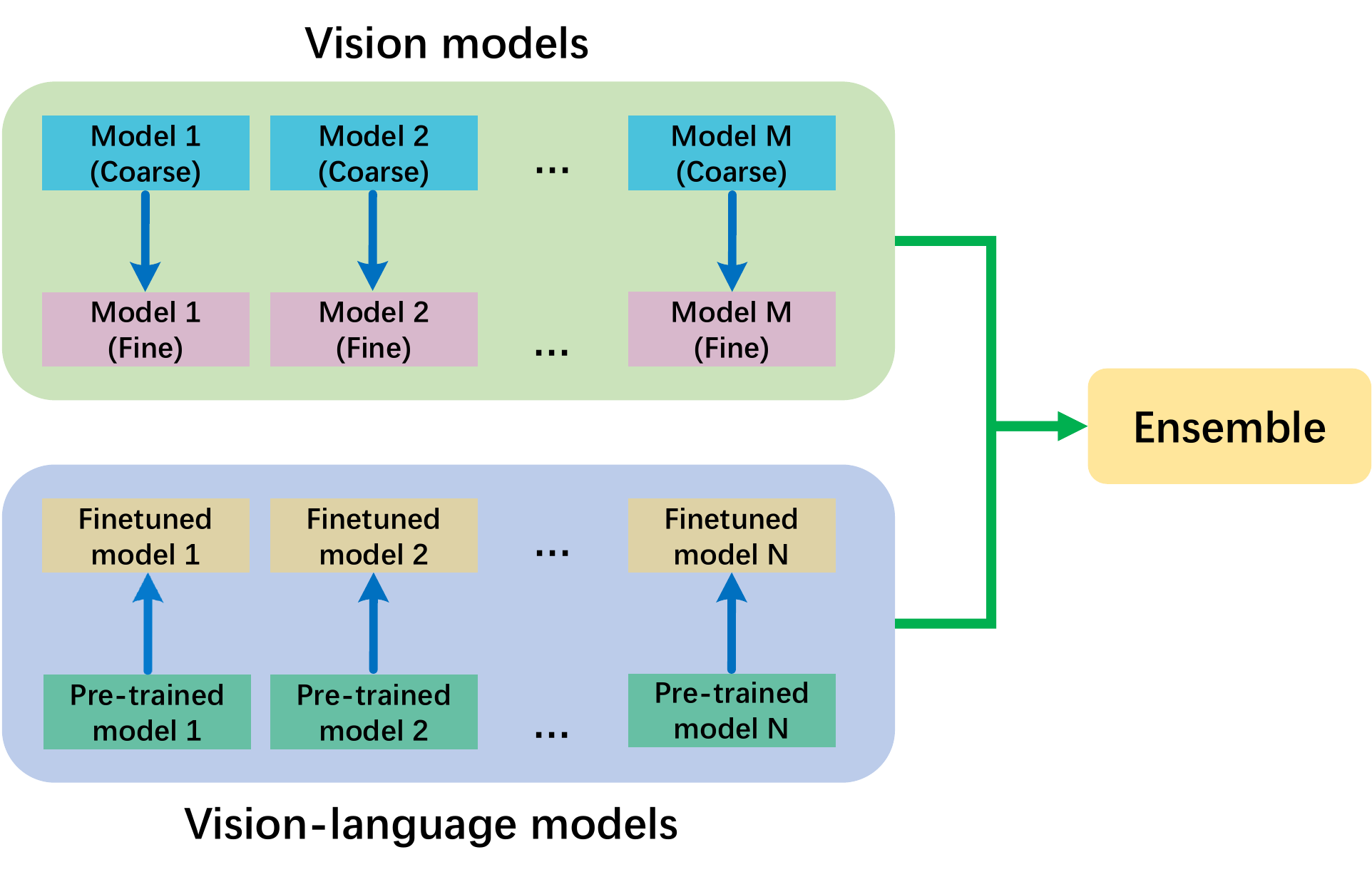}
\vspace*{2mm}
\caption{The designed V$^2$L approach. We leverage vision models and vision-language models into large-scale product retrieval. The final submission is obtained by ensembling different models' results from two types of approaches.} 
\vspace*{-4mm}
\label{frame_1}
\end{figure}

 
The proposed approach consists of two parts, i.e. vision models, and vision-language models. For vision models, we perform a coarse-to-fine training strategy. First, backbones are trained with the provided $1,000$ categories labels (without ImageNet pre-training). Then we adapt the strong self-supervised baseline in \cite{wang2021bag} to accommodate the product retrieval task. We choose many vision-language models, which are pre-trained and with suitable licenses (both code and the models themselves), and then finetune them using the product images and corresponding titles. That is how we perform multimodal learning. The final submission is obtained by ensembling different models' results from two types of approaches. The illustration of the proposed V$^2$L is shown in Fig. \ref{frame_1}.
In summary, the main contributions of this paper are:
\begin{enumerate}
 \item The paper demonstrates the effectiveness of vision-language models in product retrieval tasks for the first time.
  \item We explore the performance threshold of current product retrieval tasks by using unimaginable resources.
 \item The proposed V$^2$L approach ranks first in eBay eProduct Visual Search Challenge (FGVC9).
\end{enumerate}

\section{Proposed Method}
In this section, we introduce each important component in the proposed V$^2$L. We will first introduce our vision models and then show the selected vision-language models. Finally, to promote reproducibility, all the used tricks are listed.

\subsection{Vision Models}
We basically follow the training process in the strong baseline from \cite{wang2021bag}. The changes are listed below.
\subsubsection{Two-stage Training}
We perform coarse-to-fine training. First, the randomly initialized backbones are trained using the $1,000$ coarse labels. Note that, we do NOT use the ImageNet-pre-trained backbones due to two issues: (1) The open-source codebases often do NOT include an explicit license for their released pre-trained models. We do NOT want to involve in the license issue in the final checking. (2) It seems that the scale of eBay training dataset is large enough for training from scratch. Then, using the strong baseline \cite{wang2021bag}, the second stage training is performed. We use $8$ and $4$ Nvidia Tesla A100 GPUs for the two stages of training, respectively.
\subsubsection{Pseudo-label Generation}
First, we randomly select $100,000$ training images from the training set. Through augmentation, each training image forms its only class. A model is trained on this $100,000$ classes. After training, features extracted by the model can used for clustering. In this way, the training images are clustered. Note that, we only keep clusters with high confidence, \textit{i.e.} each cluster only contains less than $10$ images, and thus we only use about $1/5$ training images in the second stage training.
Specifically, after the clustering, we get $87,125$ classes. Then we randomly select $100K-87,125= 12,875$ images from the training dataset. That means besides the $87,125$ clusters, we have $12,875$ extra classes (each class with one image). As a result, we have $246,926$ images from the $87,125$ clusters ($2.8$ images for each cluster on average). Then, the $12,875$ extra classes bring $12,875$ extra images. Therefore, finally we have $246,926 + 12,875 = 255,801$ images.

\subsubsection{Stronger Backbones}
Instead of using the backbones in \cite{wang2021bag}, in this competition, we use much stronger backbones with suitable licenses: ResNeSt \cite{zhang2020resnest} (\href{https://github.com/zhanghang1989/ResNeSt/blob/master/LICENSE}{license}), ResNeXt \cite{xie2016aggregated} (\href{https://github.com/prlz77/ResNeXt.pytorch/blob/master/LICENSE}{license}), CotNet \cite{cotnet} (\href{https://github.com/JDAI-CV/CoTNet/blob/master/LICENSE}{license}), HS-ResNet \cite{yuan2020hs} (\href{https://github.com/PaddlePaddle/PaddleClas/blob/release/2.4/LICENSE}{license}), NAT \cite{hassani2022neighborhood} (\href{https://github.com/SHI-Labs/Neighborhood-Attention-Transformer/blob/main/LICENSE}{license}), ViT (from BEIT \cite{bao2021beit}) (\href{https://github.com/microsoft/unilm/blob/master/LICENSE}{license}), and ViT (from SimMIM \cite{xie2021simmim}) (\href{https://github.com/microsoft/SimMIM/blob/main/LICENSE}{license}). The stronger backbones perform much better.
\subsubsection{Simpler Augmentations}
The augmentations used in \cite{wang2021bag} are designed for image copy detection, and thus they are too complexed. For product retrieval tasks, we only keep RandomResizedCrop, RandomRotation, RandomPerspectiveChange, RandomPadding, RandomImageUnderlay, RandomImageOverlay, RandomLightChange, RandomHorizontalFlip, RandomVerticalFlip, and GaussianBlur. For more details, please refer to the original paper \cite{wang2021bag}.
\subsubsection{Higher Resolution}
In image copy detection, $256 \times 256$ resolution is the best. For the product retrieval task, we find that $512 \times 512$ resolution performs best. Though higher resolution brings performance improvement, it also brings much computational burden to the training and reference process.
\subsubsection{Loss}
To be computationally efficient, we replace the $8192$-dim features \cite{wang2021bag} with commonly-used $2048$-dim features. Moreover, instead of combining cross-entropy loss and triplet loss \cite{hermans2017defense}, we use a single CosFace \cite{wang2018cosface}.
\subsubsection{Exponential Moving Average}
Exponential Moving Average (EMA) is used to stabilize the training process of deep metric learning. It brings consistent performance improvement without increasing any burden.
\begin{figure*}[t]
\centering 
\includegraphics[width=0.8\textwidth]{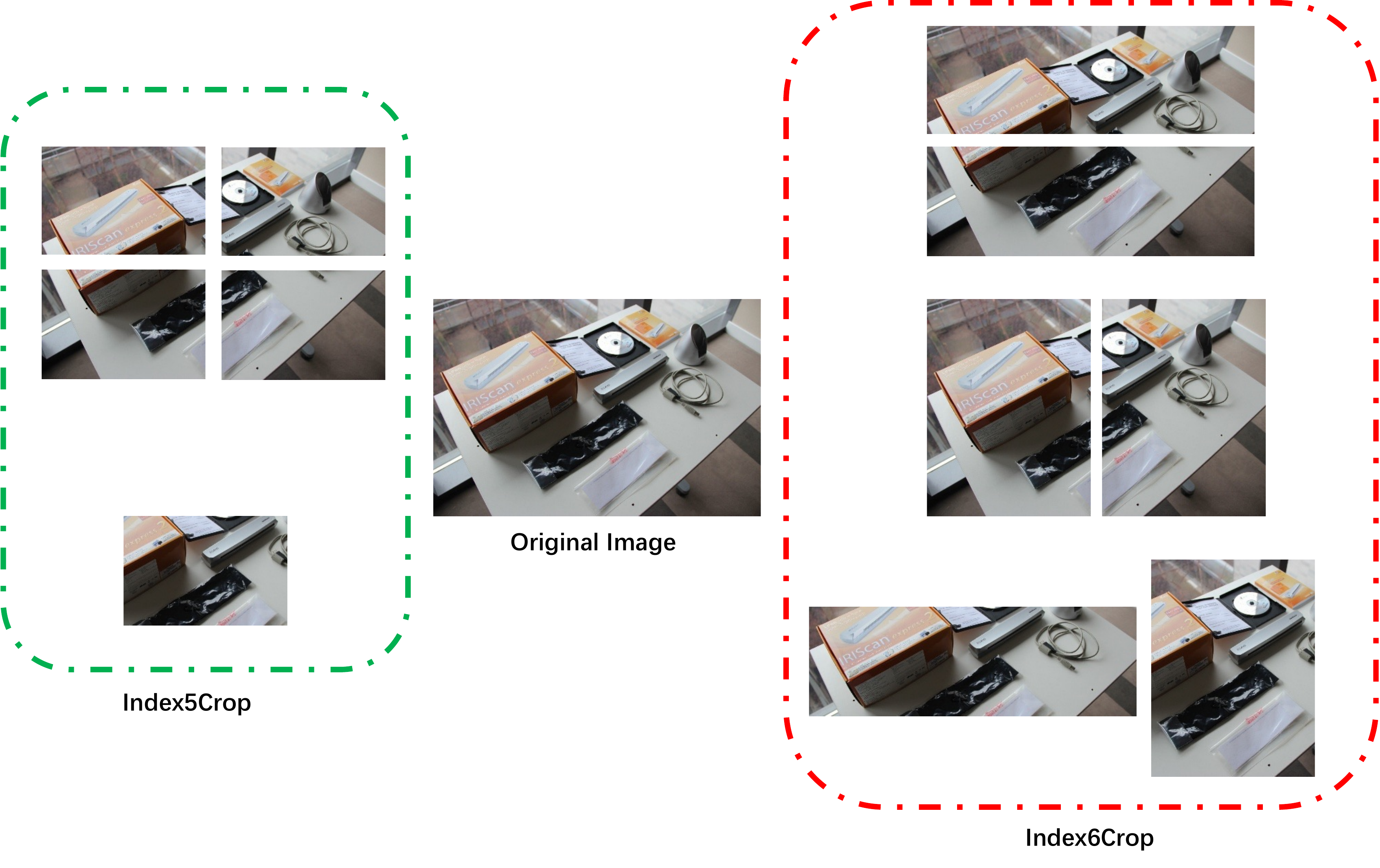}
\vspace*{2mm}
\caption{The used Index5Crop and Index6Crop. The index images are splited into $5$ or $6$ parts to be matched locally. Though it brings $5 \times$ or $6 \times$ computational burden, it significantly alleviates the side effect of background.} 
\label{index_crop}
\end{figure*}

\subsection{Vision-language Models}
In this section, we introduce the selected vision-language models and how we use that. Due to the license issue, we only choose BLIP \cite{li2022blip} (\href{https://github.com/salesforce/BLIP/issues/52}{license}), ALBEF \cite{ALBEF} (\href{https://github.com/salesforce/ALBEF/issues/80}{license}), XVLM \cite{xvlm} (\href{https://github.com/zengyan-97/X-VLM/issues/8}{license}), METER \cite{dou2022meter} (\href{https://github.com/zdou0830/METER/issues/17}{license}), and SLIP \cite{mu2021slip} (\href{https://github.com/facebookresearch/SLIP/issues/17}{license}). We get the suitable licenses for both the code and the released models. We do NOT use the famous CLIP \cite{pmlr-v139-radford21a} because we cannot get the license for the pre-trained models. \par
The finetuning process is relatively easy. Because each algorithm takes the nature language and the image as input, we just use the product title as the description of one product image. After training, the image encoder is used to extract the features of query and reference images. Though the use of the vision-language models is easy, it provides a totally different prospective to the product retrieval task.

\subsection{Ensemble Methods}
\subsubsection{Maximum Ensemble}
The ensemble method in \cite{wang2021d} is proved to be effective in the product retrieval task.
\subsubsection{Voting Ensemble}
Each model returns a ranking list, and we fetch the top-10 images. Many models vote for the final ranking list. It is similar (or same) with last year's ``Multiple networks rankings''.
\subsection{Other Tricks}
\subsubsection{Global-local Matching}
We split reference (index) images to match locally (The query images are kept). We name the used two strategies as Index5Crop and Index6Crop. They are illustrated in Fig . \ref{index_crop}. Note that we only re-train $3$ different methods to perform Index5Crop ($2$) and Index6Crop ($1$). The gained results are absorbed to final ranking by voting ensemble.

\subsubsection{Re-ranking}
The k-reciprocal re-ranking \cite{zhong2017re} is proved to be very useful for image retrieval tasks. However, it is not straightforward to apply it to millions of reference images. We distribute it with about $100$ CPUs. The advantage is the speed is much faster, and the memory is enough. However, we cannot promise $100$ CPUs work normally at the same time. Therefore, there may be some queries missing during the re-ranking process. We argue that because we ensemble a lot of models, some queries missing do not bring serious influence to the final result.
\subsubsection{Multi-scale Testing}
Because $512 \times 512$ resolution performs best, in the multi-scale testing, we use $400 \times 400$, $512 \times 512$, and $600 \times 600$ resolutions.

\begin{figure*}[t]
\centering 
\includegraphics[width=1\textwidth]{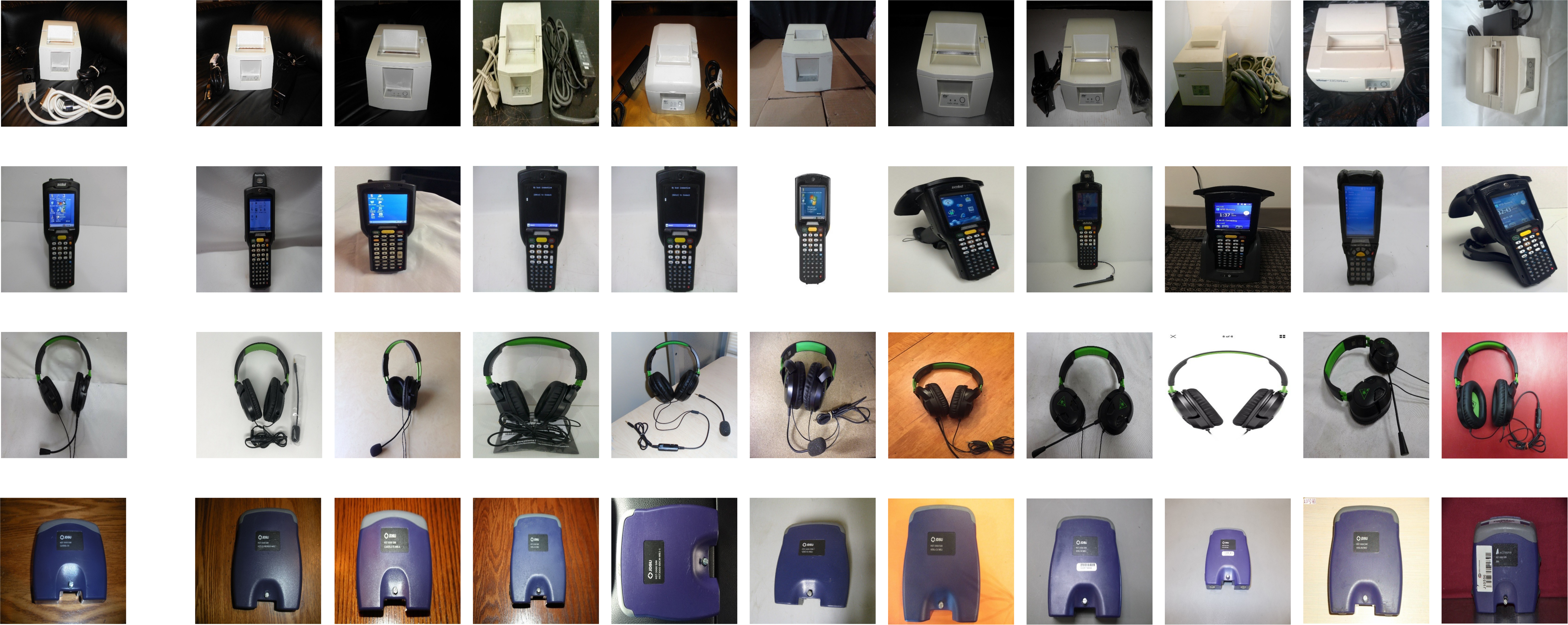}
\vspace*{2mm}
\caption{The visualization of the ranking list. In each row, the first image is the query image, and the following $10$ images are the index images.} 
\label{visual}
\end{figure*}

\section{Ineffective Methods}
We have tried a lot of methods in this competition, and only some of them work in the final ensemble period. The listed ineffective methods are representative and seem to work.
\subsection{Title for Pseudo-labels}
Last year's winner finds using the product titles to create pseudo-labels useful. We find that, when the MAR@10 is relatively low (less than $70$), this approach is effective regardless of using a single model or ensemble. However, when the MAR@10 approaches $75$, regardless of using BERT \cite{devlin2018bert} or word2vec \cite{mikolov2013efficient} to generate sentence embeddings, this approach cannot contribute to the final ensemble result.
\subsection{Multiple Layers Features}
It is reasonable that deep features of two images, which have subtle difference, may be indistinguishable. However, both training models with shallow features and directly combining shallow features with deep features do not work.
\subsection{Multiple Augmented Queries}
Performing test-time augmentation (TTA) is a common practice for image retrieval. However, we find performing TTA (\textit{e.g.} center cropping, flipping) on query images do not bring performance improvement.
\subsection{Query Expansion}
The standard query expansion (QE) \cite{chum2011total} and $\alpha$ query expansion ($\alpha$QE) \cite{radenovic2018fine} are widely used for image retrieval. However, both of them do not work in our approach.
\subsection{Background Removal}
We try to use a model to remove the background of all the query and index images. However, the performance decreases. It may because the background removal algorithm removes some discriminative matching information.
\section{Experiments}
To prove the superiority of the V$^2$L,  we compare the proposed model with state-of-the-art methods from the leaderboard in the testing phase. The comparison results are shown in Table \ref{sota}. We ranks first among all the competitors. We discover that (1) Our solution is much higher than other competitors'. (2) There is an obvious performance gap between top-3 and others' scores. Also, we visualize some results of our final ranking list. They are shown in Fig. \ref{visual}.
\begin{table}
  \caption{Comparison with state-of-the-art methods from the leaderboard in the testing phase. By ensembling about $20$ models, we rank first in the testing phase of the competition.}
  \vspace*{2mm}
\small
  \begin{tabularx}{\hsize}{|p{2.5cm}|Y|}
    \hline
    \multicolumn{1}{|c|}{\multirow{1}{*}{Team}} &
    \multicolumn{1}{c|}{MAR@10}  \\
    \hline\hline
    \textbf{CVALLStar}  & $\textbf{0.762274}$  \\
    Involution King  & $0.715327$   \\
    USTC-IAT-United  & $0.701012$  \\
    fgvc9  & $0.659382$  \\
    ums\_v1  & $0.618245$  \\
    OPPO Research& $0.547500$  \\
    AI\_VIS & $0.532808$  \\
    ...&... \\
    \hline
  \end{tabularx}

  \label{sota}
\end{table}


\section{Conclusion}
In this paper, we introduce our winning solution to the eBay eProduct Visual Search Challenge (FGVC9) at CVPR'22. The proposed V$^2$L combines vision models and vision-language models. Experiments show that the combination and the ensemble are the key to our winning.

{\small
\bibliographystyle{ieee_fullname}
\bibliography{egbib}

\begin{thebibliography}{10}\itemsep=-1pt

\bibitem{bai2020products}
Yalong Bai, Yuxiang Chen, Wei Yu, Linfang Wang, and Wei Zhang.
\newblock Products-10k: A large-scale product recognition dataset.
\newblock {\em arXiv preprint arXiv:2008.10545}, 2020.

\bibitem{bao2021beit}
Hangbo Bao, Li Dong, and Furu Wei.
\newblock Beit: Bert pre-training of image transformers.
\newblock {\em arXiv preprint arXiv:2106.08254}, 2021.

\bibitem{Le_2020_ECCV}
Lele Cheng, Xiangzeng Zhou, Liming Zhao, Dangwei Li, Hong Shang, Yun Zheng, Pan
  Pan, and Yinghui Xu.
\newblock Weakly supervised learning with side information for noisy labeled
  images.
\newblock In {\em The European Conference on Computer Vision (ECCV)}, August
  2020.

\bibitem{chum2011total}
Ond{\v{r}}ej Chum, Andrej Mikulik, Michal Perdoch, and Ji{\v{r}}{\'\i} Matas.
\newblock Total recall ii: Query expansion revisited.
\newblock In {\em CVPR 2011}, pages 889--896. IEEE, 2011.

\bibitem{devlin2018bert}
Jacob Devlin, Ming-Wei Chang, Kenton Lee, and Kristina Toutanova.
\newblock Bert: Pre-training of deep bidirectional transformers for language
  understanding.
\newblock {\em arXiv preprint arXiv:1810.04805}, 2018.

\bibitem{dou2022meter}
Zi-Yi Dou, Yichong Xu, Zhe Gan, Jianfeng Wang, Shuohang Wang, Lijuan Wang,
  Chenguang Zhu, Pengchuan Zhang, Lu Yuan, Nanyun Peng, Zicheng Liu, and
  Michael Zeng.
\newblock An empirical study of training end-to-end vision-and-language
  transformers.
\newblock In {\em Conference on Computer Vision and Pattern Recognition
  (CVPR)}, 2022.

\bibitem{hassani2022neighborhood}
Ali Hassani, Steven Walton, Jiachen Li, Shen Li, and Humphrey Shi.
\newblock Neighborhood attention transformer.
\newblock 2022.

\bibitem{hermans2017defense}
Alexander Hermans, Lucas Beyer, and Bastian Leibe.
\newblock In defense of the triplet loss for person re-identification.
\newblock {\em arXiv preprint arXiv:1703.07737}, 2017.

\bibitem{li2022blip}
Junnan Li, Dongxu Li, Caiming Xiong, and Steven Hoi.
\newblock Blip: Bootstrapping language-image pre-training for unified
  vision-language understanding and generation.
\newblock In {\em ICML}, 2022.

\bibitem{ALBEF}
Junnan Li, Ramprasaath~R. Selvaraju, Akhilesh~Deepak Gotmare, Shafiq Joty,
  Caiming Xiong, and Steven Hoi.
\newblock Align before fuse: Vision and language representation learning with
  momentum distillation.
\newblock In {\em NeurIPS}, 2021.

\bibitem{cotnet}
Yehao Li, Ting Yao, Yingwei Pan, and Tao Mei.
\newblock Contextual transformer networks for visual recognition.
\newblock {\em arXiv preprint arXiv:2107.12292}, 2021.

\bibitem{mikolov2013efficient}
Tomas Mikolov, Kai Chen, Greg Corrado, and Jeffrey Dean.
\newblock Efficient estimation of word representations in vector space.
\newblock {\em arXiv preprint arXiv:1301.3781}, 2013.

\bibitem{mu2021slip}
Norman Mu, Alexander Kirillov, David Wagner, and Saining Xie.
\newblock Slip: Self-supervision meets language-image pre-training.
\newblock {\em arXiv preprint arXiv:2112.12750}, 2021.

\bibitem{radenovic2018fine}
Filip Radenovi{\'c}, Giorgos Tolias, and Ond{\v{r}}ej Chum.
\newblock Fine-tuning cnn image retrieval with no human annotation.
\newblock {\em IEEE transactions on pattern analysis and machine intelligence},
  41(7):1655--1668, 2018.

\bibitem{pmlr-v139-radford21a}
Alec Radford, Jong~Wook Kim, Chris Hallacy, Aditya Ramesh, Gabriel Goh,
  Sandhini Agarwal, Girish Sastry, Amanda Askell, Pamela Mishkin, Jack Clark,
  Gretchen Krueger, and Ilya Sutskever.
\newblock Learning transferable visual models from natural language
  supervision.
\newblock In Marina Meila and Tong Zhang, editors, {\em Proceedings of the 38th
  International Conference on Machine Learning}, volume 139 of {\em Proceedings
  of Machine Learning Research}, pages 8748--8763. PMLR, 18--24 Jul 2021.

\bibitem{wang2018cosface}
Hao Wang, Yitong Wang, Zheng Zhou, Xing Ji, Dihong Gong, Jingchao Zhou, Zhifeng
  Li, and Wei Liu.
\newblock Cosface: Large margin cosine loss for deep face recognition.
\newblock In {\em Proceedings of the IEEE conference on computer vision and
  pattern recognition}, pages 5265--5274, 2018.

\bibitem{wang2021d}
Wenhao Wang, Yifan Sun, Weipu Zhang, and Yi Yang.
\newblock D\^{} 2lv: A data-driven and local-verification approach for image
  copy detection.
\newblock {\em arXiv preprint arXiv:2111.07090}, 2021.

\bibitem{wang2021bag}
Wenhao Wang, Weipu Zhang, Yifan Sun, and Yi Yang.
\newblock Bag of tricks and a strong baseline for image copy detection.
\newblock {\em arXiv preprint arXiv:2111.08004}, 2021.

\bibitem{xie2016aggregated}
Saining Xie, Ross Girshick, Piotr Doll{\'a}r, Zhuowen Tu, and Kaiming He.
\newblock Aggregated residual transformations for deep neural networks.
\newblock {\em arXiv preprint arXiv:1611.05431}, 2016.

\bibitem{xie2021simmim}
Zhenda Xie, Zheng Zhang, Yue Cao, Yutong Lin, Jianmin Bao, Zhuliang Yao, Qi
  Dai, and Han Hu.
\newblock Simmim: A simple framework for masked image modeling.
\newblock {\em arXiv preprint arXiv:2111.09886}, 2021.

\bibitem{yuan2021eproduct}
Jiangbo Yuan, An-Ti Chiang, Wen Tang, and Antonio Haro.
\newblock eproduct: A million-scale visual search benchmark to address product
  recognition challenges.
\newblock {\em arXiv preprint arXiv:2107.05856}, 2021.

\bibitem{yuan2020hs}
Pengcheng Yuan, Shufei Lin, Cheng Cui, Yuning Du, Ruoyu Guo, Dongliang He,
  Errui Ding, and Shumin Han.
\newblock Hs-resnet: Hierarchical-split block on convolutional neural network.
\newblock {\em arXiv preprint arXiv:2010.07621}, 2020.

\bibitem{xvlm}
Yan Zeng, Xinsong Zhang, and Hang Li.
\newblock Multi-grained vision language pre-training: Aligning texts with
  visual concepts.
\newblock {\em arXiv preprint arXiv:2111.08276}, 2021.

\bibitem{zhang2020resnest}
Hang Zhang, Chongruo Wu, Zhongyue Zhang, Yi Zhu, Zhi Zhang, Haibin Lin, Yue
  Sun, Tong He, Jonas Muller, R. Manmatha, Mu Li, and Alexander Smola.
\newblock Resnest: Split-attention networks.
\newblock {\em arXiv preprint arXiv:2004.08955}, 2020.

\bibitem{zhong2017re}
Zhun Zhong, Liang Zheng, Donglin Cao, and Shaozi Li.
\newblock Re-ranking person re-identification with k-reciprocal encoding.
\newblock In {\em Proceedings of the IEEE conference on computer vision and
  pattern recognition}, pages 1318--1327, 2017.

\end{thebibliography}
}

\end{document}